\documentclass[11pt,a4paper]{article}
\usepackage[hyperref]{naaclhlt2018}
\usepackage{times}
\usepackage{latexsym}
\usepackage{graphicx}
\usepackage{url}
\usepackage{natbib}
\usepackage{amsmath}
\usepackage{booktabs}
\usepackage{mathtools}
\usepackage{multirow}
\usepackage{subfig}
\usepackage{amssymb}
\usepackage{tikz}
\usetikzlibrary{arrows,shapes,snakes,automata,backgrounds,petri,positioning}

\DeclareCaptionFont{10pt}{\fontsize{10pt}{12pt}\selectfont}
\aclfinalcopy 

\newcommand*\samethanks[1][\value{footnote}]{\footnotemark[#1]}

\title{Learning beyond datasets: Knowledge Graph Augmented Neural Networks for Natural language Processing }

\author{
	Annervaz K M\thanks{equal contribution}  \\
  Indian Institute of Science, \\
 Accenture Technology Labs \\ 
 annervaz@iisc.ac.in \\ \And
  Somnath Basu Roy Chowdhury\samethanks[1] \thanks{Main work done during internship at Accenture Technology Labs}\\
  IIT Kharagpur \\
  brcsomnath@ee.iitkgp.ernet.in \\ \And
	Ambedkar Dukkipati \\
	Indian Institute of Science \\ 
    ambedkar@iisc.ac.in }

\date{}

\begin{document}
\maketitle
\begin{abstract}
Machine Learning has been the quintessential solution for many AI problems, but learning models are heavily dependent on specific training data. Some learning models can be incorporated with prior knowledge using a Bayesian setup, but these learning models do not have the ability to access any organized world knowledge on demand.  In this work, we propose to enhance learning  models with \textit{world knowledge} in the form of Knowledge Graph (KG) fact triples for Natural Language Processing (NLP) tasks. Our aim is to develop a deep learning model that can extract relevant prior support facts from \textit{knowledge graphs} depending on the task using attention mechanism. We introduce a \textit{convolution-based model} for learning representations of knowledge graph entity and relation clusters in order to reduce the attention space. We show that the
proposed method is highly scalable to the amount of prior information that has to be processed and can be applied to any generic NLP task. Using this method we show significant improvement in performance for text classification with 20Newsgroups (News20) \& DBPedia datasets, and natural language inference with Stanford Natural Language Inference (SNLI) dataset. We also demonstrate that a deep learning model can be trained with substantially less amount of labeled training data, when it has access to organized world knowledge in the form of a knowledge base. 
\end{abstract}

\section{Introduction}

Today, machine learning is centered around algorithms that can be trained on available task-specific labeled and unlabeled training samples. Although learning paradigms like Transfer Learning~\cite{pan2010survey} attempt to incorporate knowledge from one task into another, these techniques are limited in scalability and are specific to the task at hand.  On the other hand, humans have the intrinsic ability to elicit required past knowledge from the world on demand and infuse it with newly learned concepts to solve problems.

The question that we address in this paper is the following: Is it possible to develop learning models that can be trained in a way that it is able to infuse a general body of world knowledge for prediction apart from learning based on training data?  

\begin{figure}[h!]
\begin{tikzpicture}[node distance=1cm,>=stealth',bend angle=45,auto]
	\tikzstyle{place}=[circle,thin,draw=black!75,fill=white!20,minimum size=10mm]
	\tikzstyle{transition}=[rectangle,thin,draw=black!75, fill=white!20,minimum size=12mm]
	\tikzstyle{empty}=[rectangle,thin,draw=white!75, fill=white!20,minimum size=8mm]
	\begin{scope}
	\node[transition](r1) [draw, text width=2.3cm,  align=center, text centered]{\textit{World-Knowledge Base}};
	\node [place] (w1) [right of=r1, xshift=1.5cm] {$\mathcal{X}$}
	 edge [post] (r1);
	\node [empty] (w2) [below of=w1] {$+$};
	\node [place] (w3) [below of=w2] {$\mathcal{X}_w$}
	edge [pre, bend left] (r1);
	\node [place] (w4) [right of=w2, xshift=0.75cm] {$\mathcal{X}'$}
	edge [pre] (w2);
	\node [place] (w5) [right of=w4, xshift=0.75cm] {$\mathcal{Y}$}
	edge [pre] node[above]{$f$} (w4);
	\end{scope}
\end{tikzpicture}
\caption{The Basic Idea: $\mathcal{X}$ is the feature input and $\mathcal{Y}$ is the prediction. The relevant world knowledge for the task $\mathcal{X}_w$, is retrieved and augmented with the feature input before making the final prediction}
\label{fig:intuition}
\end{figure}
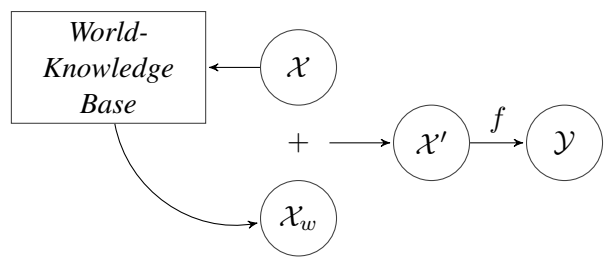 


By world knowledge, we mean structured general purpose knowledge that need not be domain specific. Knowledge Graphs~\cite{nickel2016review} 
 are a popular source of such structured world knowledge. Knowledge Graphs represent information in the form of fact triplets, consisting of a subject entity, relation and object entity (example: $<$\textit{Italy, capital, Rome}$>$). The entities represent the nodes of the graph and their relations act as edges. A fact triple $($subject entity, relation, object relation$)$ is represented as $(h,r,t)$.  Practical knowledge bases congregate information from secondary databases or extract facts from unstructured text using various statistical learning mechanisms, examples of such systems are NELL~\cite{mitchell2015never} and DeepDive~\cite{niu2012deepdive}. There are human created knowledge bases as well, like Freebase (FB15k)~\cite{bollacker2008freebase} and WordNet~\cite{miller1990introduction}. The knowledge present in 
these knowledge bases includes common knowledge and partially covers common-sense knowledge and domain knowledge~\cite{song2017machine}. Knowledge Graphs and Knowledge Bases are conceptually equivalent for our purpose and we will use the name interchangeably in this paper.  

We illustrate the significance of world knowledge using a few examples.  For the example of a Natural Language Inference (NLI) problem~\cite{maccartney2009natural}, consider the two following statements, A: \texttt{The couple is walking on the sea shore} and B: \texttt{The man and woman are wide awake}. Here, for a learning model to infer B from A, it should have access to the common knowledge that ``\textit{The man and woman} and \textit{The couple} means the same'' since this information may not be specific for a particular inference. Further, it is not possible for a model to learn all such correlations from just the labeled training data available for the task.

Consider another example of classifying the news snippet, \texttt{Donald Trump offered his condolences towards the hurricane victims and their families in Texas}. We cannot classify it as a political news unless we know the facts $<$\textit{Donald Trump, president, United States}$>$ and $<$\textit{Texas, state, United States}$>$. We posit that machine learning models, apart from training them on data with the ground-truth can also be trained to fetch relevant information from structured knowledge bases in order to enhance their performance.  

In this work, we propose a deep learning model that can extract relevant support facts on demand from a knowledge 
base~\cite{mitchell2015never} and incorporate it in the feature space along with the features learned from the training data (shown in Figure \ref{fig:intuition}). This is a challenging task, as knowledge bases typically have millions of fact triples. Our proposed model involves a deep learning mechanism to jointly model this look-up scheme along with the task specific training of the model. The look-up mechanism and model is generic enough so that it can be augmented to any task specific learning model to boost the learning performance. In this paper, we have established superior performance of the proposed KG-augmented models over vanilla model on text classification and natural language inference.

Although there is a plethora of work on knowledge graph representation~\cite{nickel2016review}~\cite{mitchell2015never}~\cite{niu2012deepdive} from natural language text, no attempt to augment learning models with knowledge graph information have been done. To the best of our knowledge this is the first attempt to incorporate world knowledge from a knowledge base for learning models.  



\section{Knowledge Graph Representations}

Knowledge Graph entities/relations need to be encoded into a numerical representation for processing. 
 Before describing the model, we provide a brief overview of graph encoding techniques. Various KG embedding techniques can be classified at a high level into: \textit{Structure-based embeddings} and \textit{Semantically-enriched embeddings}.

\textbf{Structure-based embeddings}: TransE~\cite{bordes2013translating} is the introductory work on knowledge graph representation, which translated subject entity to object entity using one-dimensional relation vector $(h+r=t)$. Variants of the TransE \cite{bordes2013translating} model uses translation of the entity vectors over relation specific subspaces. TransH \cite{wang2014knowledge} introduced the relation-specific hyperplane to translate the entities. Similar work utilizing only the structure of the graph include  ManifoldE \cite{xiao2015one}, TransG \cite{xiao2015transg}, TransD \cite{ji2015knowledge}, TransM \cite{fan2014transition}, HolE \cite{nickel2016holographic} and ProjE \cite{shi2017proje}.

\textbf{Semantically-enriched embeddings}: These embedding techniques learn to represent entities/relations of the KG along with its semantic information.  Neural Tensor Network(NTN) \cite{socher2013reasoning} was the pioneering work in this field which initialized entity vectors with the average word embeddings followed by tensor-based operations. Recent works involving this idea are ``Joint Alignment" \cite{zhong2015aligning} and SSP \cite{xiao2017ssp}. DKRL \cite{xie2016representation} is a KG representation technique which also takes into account the descriptive nature of text keeping the simple structure of TransE model. Pre-trained word2vec~\cite{mikolov2013efficient} is used to form the entity representation by passing through a Convolutional Neural Network (CNN)~\cite{kim2014convolutional} architecture constraining the relationships to hold. 

In our experiments we have used the DKRL \cite{xie2016representation} encoding scheme as it emphasizes on the semantic description of the text. Moreover, DKRL fundamentally uses TransE \cite{bordes2013translating} method for encoding structural information. Therefore, we can retrieve relevant entities \& relation and obtain the complete the fact using $t=h+r$. This reduces the complexity of fact retrieval as the number of entities/relations is much less compared to the number of facts, thus making the retrieval process faster. 

\section{The Proposed Model}
\label{secmain}

\begin{figure*}[h!]
	\centering
	\includegraphics[width=0.9\textwidth, keepaspectratio]{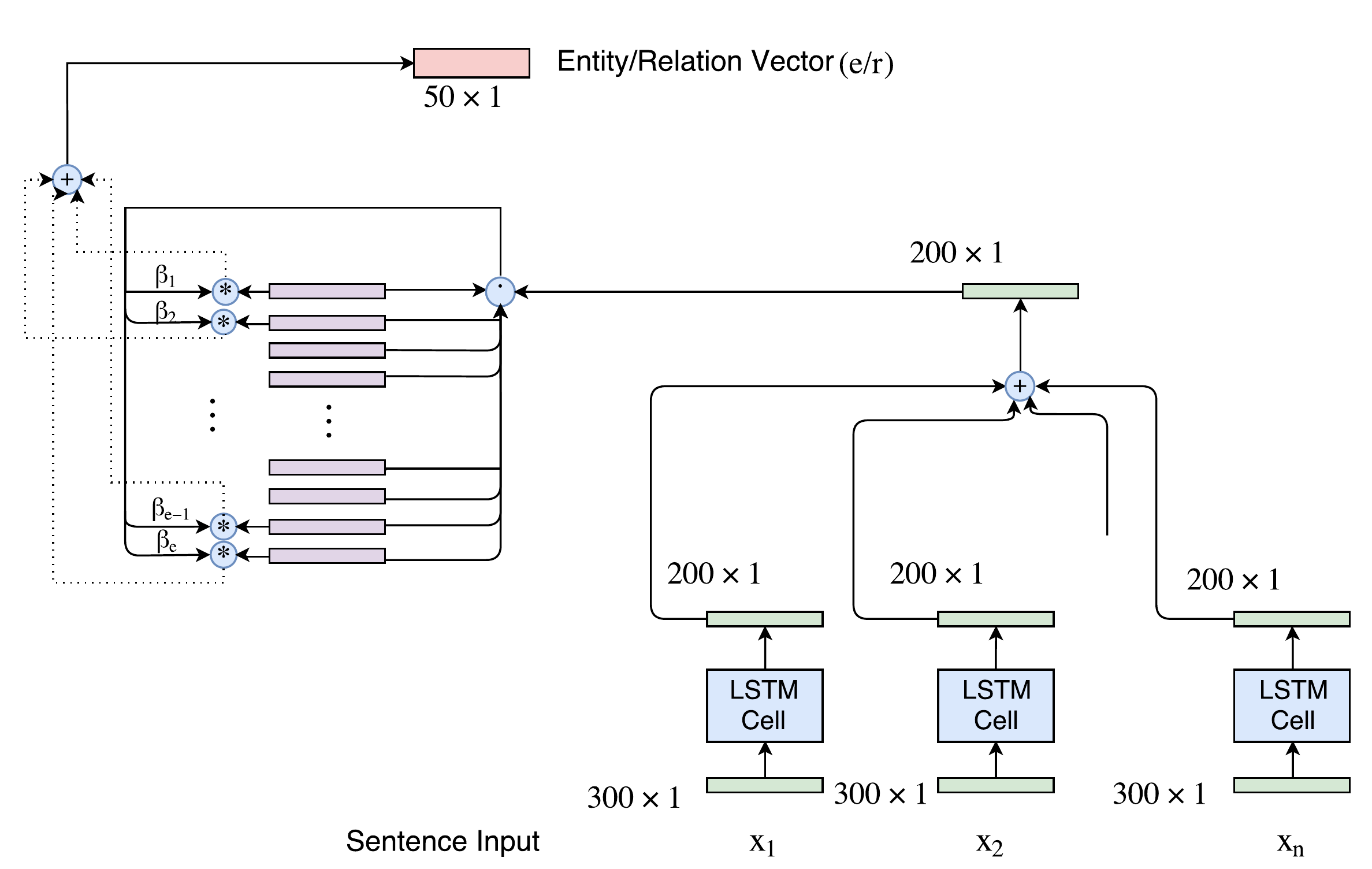}
	\caption{Vanilla Entity/Relationship Retrieval Block Diagram}
	\label{fig:operator}
\end{figure*}

Conventional supervised learning models with parameters $\Theta$, given training data $x$ and label $y$, tries to maximize the following function
$$\mathop{\mathrm{max}}\limits_{\Theta} P(y| x, \Theta)$$

The optimized parameters $\Theta$ are given as,

$$\Theta = \mathop{\mathrm{arg max}}\limits_{\Theta} \log P(y| x, \Theta)$$

\noindent In this work, we propose to augment the supervised learning process by incorporation of world knowledge features $x_w$. The world knowledge features are retrieved using the data $x$, using a separate model where, $x_w = F(x, \Theta^{(2)})$. Thus, our modified objective function can be expressed as
$$\mathop{\mathrm{max}}\limits_{\Theta} P(y| x, x_w, \Theta^{(1)})$$

where, $\Theta = \{\Theta^{(1)}, \Theta^{(2)}\}$. The optimized parameters can be obtained using the equation
$$\Theta = \mathop{\mathrm{arg max}}\limits_{\Theta} \log P(y| x, F(x, \Theta^{(2)}) ,\Theta^{(1)})$$

The subsequent sections focus on the formulation of the function $F$ which is responsible for fact triple retrieval using the data sample $x$. Here it is important to note that, we are not assuming any structural form for $P$ based on $F$. So the method is generic and applicable to augment any supervised learning setting with any form for $P$, only constraint being $P$ should be such that the error gradient can be computed with respect to $F$.  In the experiments we have used softmax using the LSTM~\cite{greff2015lstm} encodings of the input as the form for $P$. As for $F$, we use soft attention~\cite{luong2015effective,bahdanau2014neural} using the LSTM encodings of the input and appropriate representations of the fact(s). Based on the representation used for the facts, we propose two models $(a)$ Vanilla Model $(b)$ Convolution-based entity/relation cluster representation, for fact retrieval in the subsequent sections.

\subsection{Vanilla Model}

The entities and relationships of KG are encoded using DKRL, explained earlier. Let $e_i \in \mathbb{R}^m $ stand for the encoding of the entity $i$ and $r_j \in \mathbb{R}^m$ stands for $j^{\mathrm{th}}$ relationship in the KG. The input text in the form of concatenated word vectors, $\boldsymbol{x} = (x_1, x_2, \ldots, x_T)$ is first encoded using an LSTM~\cite{greff2015lstm} module as follows, $$h_t = f(x_t, h_{t-1})$$
and $$o = \frac{1}{T}{\sum_{t=1}^{T}h_t},$$ $h_t \in \mathbb{R}^n$ is the hidden state of the LSTM at time $t$, $f$ is a non-linear function and $T$ is the sequence length. Then a context vector is formed from $o$ as follows, $$C = \mathrm{ReLU}(o^TW),$$ where,  $W \in \mathbb{R}^{n \times m}$ represent the weight parameters. The same procedure is duplicated with separate LSTMs to form two seperate context vectors, one for entity retrieval ($C_{E}$) and one for relationship retrieval ($C_{R}$).

As the number of fact triples in a KG is in the order of millions in the vanilla model, we resort to generating attention over the entity and relation space separately. The fact is then formed using the retrieved entity and relation. The attention for the entity, $e_i$ using entity context vector is given by $$\alpha_{e_{i}} = \frac{\exp(C_{E}^T e_i)}{\sum\limits_{j=0}^{|E|} \exp(C_{E}^Te_j)}$$ where $|E|$ is the number of entities in the KG.

Similarly the attention for a relation vector $r_i$  is computed as $$\alpha_{r_{i}} = \frac{\exp(C_{R}^T r_i)}{\sum\limits_{j=0}^{|R|} \exp(C_{R}^Tr_j)}$$ where $|R|$ is the number of relations in the KG. The final entity and relation vector retrieval is computed by the weighted sum with the attention values of individual retrieved entity/relation vectors. $$e = \sum\limits_{i=0}^{|E|} \alpha_{e_i} e_i \hspace*{0.5cm} r = \sum\limits_{i=0}^{|R|} \alpha_{r_i} r_i$$

Figure \ref{fig:operator} shows the schematic diagram for entity/relation retrieval. After the final entity and relation vectors are computed, we look forward to completion of the fact triple. The KG embedding technique used for the experiment is DKRL which inherently uses the TransE model assumption ($h + r \approx t$). Therefore, using the subject entity and relation we form the object entity as $t = e+r$. Thus the fact triplet retrieved is $\mathcal{F} = [e, r, e+r]$, where $\mathcal{F} \in \mathbb{R}^{3m}$. This retrieved fact information is concatenated along with the context vector ($C$) of input $x$ obtained using LSTM module. The final classification label $\mathbf{y}$ is computed as follows, 
$$\mathcal{F}^{'} = \mathrm{ReLU}(\mathcal{F}^TV)$$$$\mathbf{y} = \mathrm{softmax}([\mathcal{F}^{'}:C]^TU)$$ where, $V \in \mathbb{R}^{3m \times u}$ and $U \in \mathbb{R}^{2u \times u}$ are model parameters to be learned. $\mathbf{y}$ is used to compute the cross entropy loss. We minimize this loss averaged across the training samples, to learn the various model parameters using stochastic gradient descent~\cite{stochastic-gradient-tricks}. The final prediction $\mathbf{y}$, now includes information from both dataset specific samples and world knowledge to aid in enhanced performance. While jointly training the attention mechanism tunes itself to retrieve relevant facts that are required to do the final classification. 

\subsection{Pre-training KG Retrieval}

\begin{figure*}[h!]
	\centering
	\includegraphics[width=\textwidth, keepaspectratio]{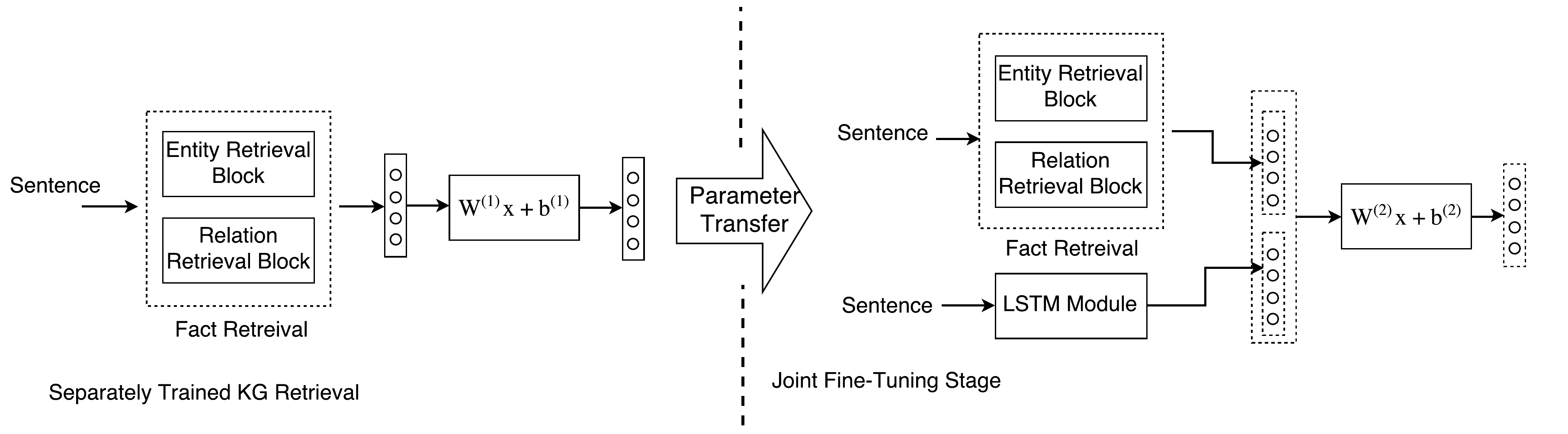}
	\caption{Separately Training Knowledge Graph Retrieval and Jointly Training the Full Model}
	\label{fig:sep_kg}
\end{figure*}

The vanilla model attends over the entire entity/relation space which is not a good approach as we observe that the  gradient for each attention value gets saturated easily. While training the classification and  retrieval module together, the model tends to ignore the KG part and gradient propagates only through the classification module. This is expected to an extent as the most pertinent information for the task at hand comes from the training samples, only background aiding information comes from KG. After few epochs of training, the KG retrieved fact always converged to a fixed vector. To overcome this problem, we attempted pre-training KG retrieval part separately. A pre-trained KG model is used to retrieve the facts and then concatenate with the classification module, while we allow error to be propagate through the pre-trained model, at the time of joint training. We infer that KG doesn't return noise and has essential information for the task as the separate KG part alone shows significant performance (59\% for News20 \& 66\% for SNLI). Figure \ref{fig:sep_kg} depicts the entire training scheme. This procedure solved the issue of gradient saturation in the KG retrieval part at the time of joint training. However, the key problem of attention mechanism having to cover a large span of entities/relation, remained.

\subsection{Convolution-based Entity and Relation Cluster Representation}

In this section, we propose a mechanism to reduce the large number of entities/relationships over which attention has to be generated in the knowledge graph. We propose to reduce the attention space by learning the representation of similar entity/relation vectors and attending over them.
\begin{figure}[h!]
	\centering
	\includegraphics[scale=0.4]{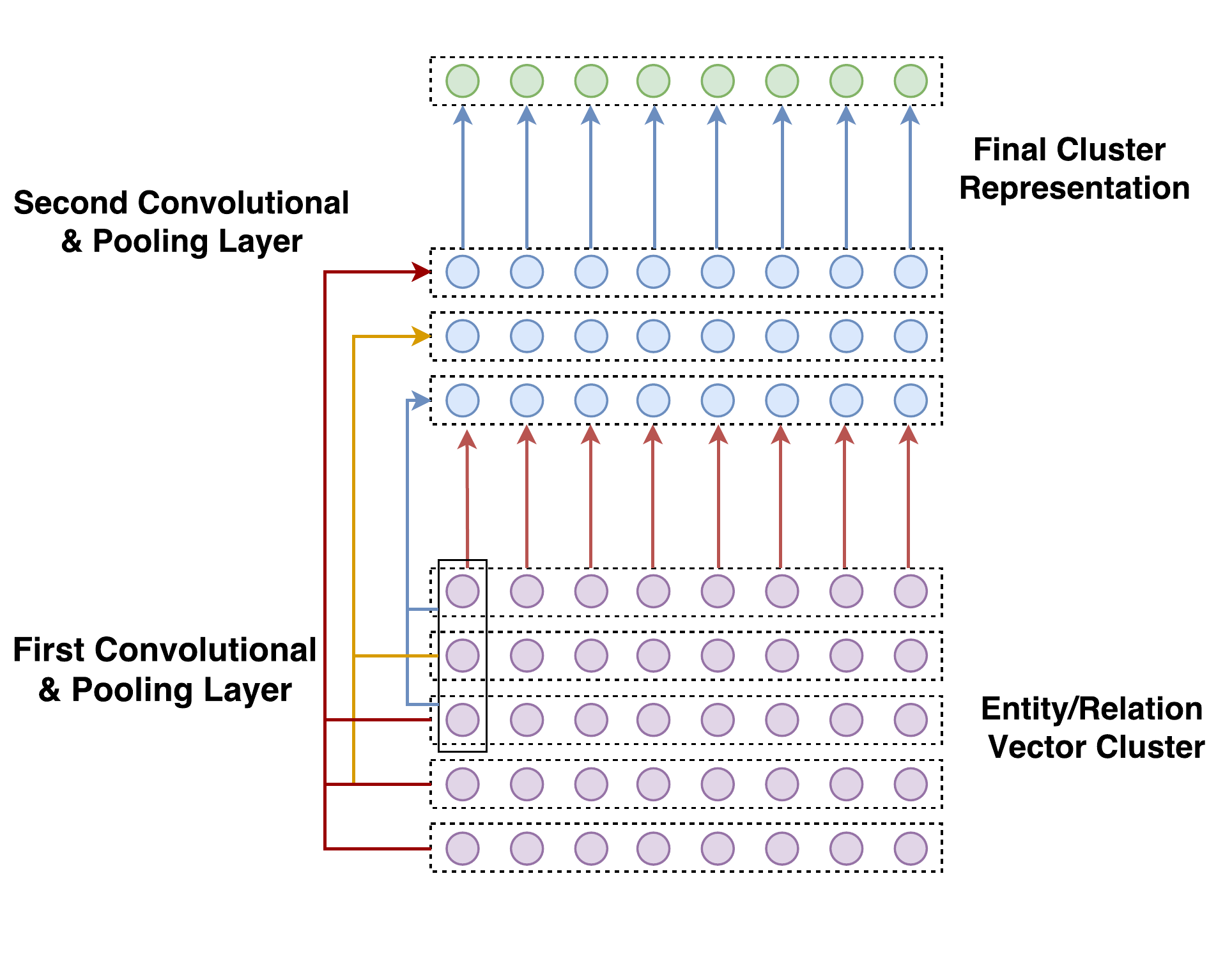}
	\caption{Convolution model cluster representation}
	\label{fig:conv_model}
\end{figure}


\noindent In order to cluster similar entity/relation vectors, we used  $k$-means clustering~\cite{bishop2006pattern} and formed $l$ clusters with equal number of entity/relation vectors in each cluster. Each of the clusters were then encoded using convolutional filters. The output of the $k$-means clustering is a sequence of entity/relation vectors $\{e_1^T, e_2^T, \cdots, e_{q}^T \}$, where $e_i \in \mathbb{R}^m$. For each cluster these vectors were stacked to form  $\mathcal{E}$ as the 2-D input to the CNN encoder, where $\mathcal{E} \in \mathbb{R}^{m \times q}$. During experimentation for finding a suitable filter shape, it was observed that using 2-D filters the model failed to converge at all. Therefore, we inferred that the latent representation of two different indices in the vector $e_i$, should not be tampered using convolution. We then resorted to use 1-D convolution filters which slide along only the  columns of $\mathcal{E}$, as shown Figure \ref{fig:conv_model}. The stride length along $y$-axis is the window length $k$. The output of the convolution layer is expressed as,
$$\mathcal{E}'(i,j) = W^T[e_{i,j}, e_{i+1,j}, \ldots, e_{i+k-1,j}]^T$$ where, $\mathcal{E}'(i,j)$ is the $(i,j)^{th}$ element of the output matrix $\mathcal{E}'$ and $W \in \mathbb{R}^{k}$ is the convolution weight filter. A pooling layer followed the convolution layer in order to reduce the parameter space, we used 1-D window only along the $y$-axis similar to the convolutional kernel mentioned above. We used a two layered convolution network with the stride length $k$ \& max-pool windows $n$ is adjusted to obtain output $E_i \in \mathbb{R}^m$, where $i$ is the cluster index. Similar procedure of clustering followed by the encoding of the cluster entities is done for relations as well. Thus both the entity and relation space were reduced to contain fewer elements, one each for each cluster. After the compact entity space $E$ and relation space $R$ is formed, we followed the same steps as earlier for forming the attention, but now the training was more effective as the gradient was propagating effectively and was not choked by the large space. As the convolution architecture is also simultaneously trained, attention mechanism was not burdened as before, to learn over the large space of entities and relations. 

Another point that needs to be mentioned here is regarding ranking/ordering items in the clusters, we have done experiments to verify the ordering does not affect the final result. We have verified this by randomly shuffling the entities/relationships in every clusters and the accuracy output remained within an error bound of $\pm 0.5\%$. In various permutations, the representations learned by the convolution operator for clusters varies, but it does not affect the overall results. Regarding the interpretation of what convolution operator learns, the operator is applied along each dimension of the entity/relationship vector, to learn a representation of the clusters. This representation includes information from relevant entities in the cluster, as the relevant entities varies across tasks, the representation learned using convolution also adapts accordingly. 
It is analogous to learning relevant features from an image, in our case the convolution layer learns the features focusing on relevant entities/relations in a cluster pertaining to the task.

\section{Experiments and Evaluations}
\label{secee}

Our experiments were designed to analyze whether a deep learning model is being improved when it has access to KG facts from a relevant source. The selection of knowledge graph has to be pertinent to the task at hand, as currently there is no single knowledge base that contains multiple kinds of information and can cater to all tasks. We illustrate with results that the performance of a deep learning model improves when it has access to relevant facts. We also illustrate that as the model learns faster with access to knowledge bases, we can train deep learning models with substantially less training data, without compromising on the accuracy. In the subsequent section we briefly describe the datasets and associated Knowledge Bases used.

\subsection*{Datasets and Relevant Knowledge Graphs}

In our experiments, we have mainly used the popular text classification dataset 20Newsgroups \cite{Lichman:2013} and the Natural Language Inference dataset, Stanford Natural Language Inference (SNLI) corpus \cite{snli:emnlp2015}.  We have also done experiments on DBPedia ontology classification dataset\footnote{\url{http://wiki.dbpedia.org/services-resources/dbpedia-data-set-2014}}, with a very strong baseline. 
These datasets are chosen as they share domain knowledge with two most popular knowledge bases, Freebase (FB15k) \cite{bollacker2008freebase} and WordNet (WN18) \cite{bordes2013translating}. The training and test size of the datasets are mentioned in Table ~\ref{tab:dataset}.

\begin{table}[h!]
	\centering
	\begin{tabular}{ c c c c}
		\hline
		Dataset & Train Size & Test Size & \# Classes\\
		\hline
		News20 & 16000 & 2000 & 20 \\ 
		SNLI & 549367 & 9824 & 3\\ 
		DBPedia & 553,000 & 70,000 & 14\\ 
		\hline
	\end{tabular}\\
	\vspace*{1mm}
	\caption{Dataset Specifications}
	\label{tab:dataset}
\end{table}

Freebase (FB15k) \cite{bollacker2008freebase} contains facts about people, places and things (contains 14904 entities, 1345 relations \& 4.9M fact triples), which is useful for text classification in 20Newsgroups \cite{Lichman:2013} dataset. On the other hand, WordNet (WN18) \cite{bordes2013translating} (has 40943 entities, 18 relations \& 1.5M fact triples) contains facts about common day-to-day things (example: furniture includes bed), which can help in inference tasks like SNLI. Both the knowledge bases are directed graphs, due to fewer number of relations WN18 the entities are more likely to be connected using the same type of relations. For experiments relating to both the datasets 20Newsgroups \& SNLI we used the standard LSTM as the classification module. As iterated earlier, our KG based fact retrieval is independent of the base model used. We show improvement in performance using the proposed models by KG fact retrieval. We use classification accuracy of the test set as our evaluation metric.


\subsection{Experimental Setup}

All experiments were carried on a Dell Precision Tower 7910 server with Quadro M5000 GPU with 8 GB of memory. The models were trained using the Adam's Optimizer~\cite{kingma2014adam} in a stochastic gradient descent~\cite{stochastic-gradient-tricks} fashion. The models were implemented using TensorFlow~\cite{tensorflow2015-whitepaper}. The relevant hyper-parameters are listed in Table \ref{tab:hyperparameters}. The word embeddings for the experiments were obtained using the pre-trained GloVe~\cite{pennington2014glove}\footnote{http://nlp.stanford.edu/data/glove.840B.300d.zip} vectors. 
For words missing in the pre-trained vectors, the local GloVe vectors which was trained on the corresponding dataset was used.
\begin{table}[h!]
	\centering
	\begin{tabular}{ c  c  c}
		\hline
		\textbf{Hyper-parameter} & News20 & SNLI\\
		\\[-1em]
		\hline
		\\[-1em]
		Batch size & 256 & 1024\\ 
		Learning rate & 0.05 & 0.05\\
		Word Vector Dimension & 300 & 300\\
		Sequence Length & 300 & 85\\
		LSTM hidden-state Dimension & 200 & 200\\
		KG Embedding Dimension & 50 & 50\\
		\# Clusters & 20 & 20\\
		\# Epochs & 20 & 20\\
		\hline
	\end{tabular}\\
	\vspace*{1mm}
	\caption{Hyper-parameters which were used in experiments for News20 \& SNLI datasets}
	\label{tab:hyperparameters}
\end{table}

\subsection{Results \& Discussion}
Table \ref{tab:result} shows the results of test accuracy of the various methods proposed on the datasets News20 \& SNLI. We observe that incorporation of KG facts using the basic vanilla model improves the performance slightly, as the retrieval model was not getting trained effectively. The convolution-based model shows significant improvement over the normal LSTM classification. While tuning the parameters of the convolution for clustered entities/relations it was observed that smaller stride length and longer max-pool window improved performance.  
For News20 dataset we show an improvement of almost 3\% and for SNLI an improvement of almost 5\%. 

The work is motivated more from the perspective of whether incorporation of world knowledge will improve any deep learning model rather than beating the state-of-the-art performance. Although LSTM is used to encode the input for the model as well as the retrieval vector, as mentioned earlier, these two modules need not be same. For encoding the input  any complex state-of-the-art model can be used. LSTM has also been used to generate the retrieval vector. For DBPedia ontology classification dataset, we have used a strong baseline of 98.6\%, and after augmenting it with KG (Freebase) using convolution based model we saw an improvement of $\sim$0.2\%. As the baseline is stronger, the improvement quantum has decreased. This is quite intuitive as complex models are self-sufficient in learning from the  data by itself and therefore  the room available for further improvement is relatively less.  The improvement as observed in the experiments is significant in weaker learning models, however it is also capable of improving stronger baselines as is evident from the results of DBPedia dataset.

\begin{table}[h]
	\centering
	\begin{tabular}{ p{3.55cm} c c }
		\hline
		\multirow{2}{*}{\textbf{Model}} & \multicolumn{2}{c}{\textbf{Accuracy}}  \\ \cline{2-3}\\[-1em]
		&News20 & SNLI\\
		\hline \\[-1em]
		Plain LSTM & 66.75\% & 68.73\%\\ 
		Vanilla KG Retrieval & 67.30\% & 69.20\% \\
		Convolution-based KG &\textbf{69.34}\% & \textbf{73.10}\%\\
		\hline
	\end{tabular}\\
	\vspace*{1mm}
	\caption{Test accuracy of approaches in News20 using FB15K \& SNLI datasets using WN18}
	\label{tab:result}
\end{table}

\subsection{Reducing Dataset Size Requirements for Training Deep Learning Models}

\begin{figure}[h!]
	\includegraphics[scale=0.8]{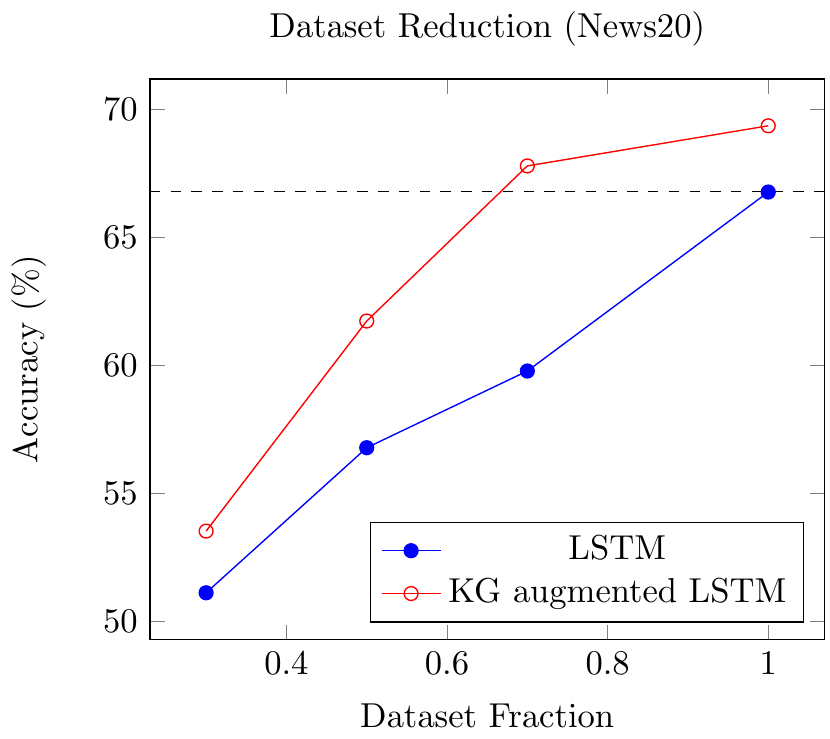}
	\caption{Accuracy Plot over dataset fractions for baseline and KG augmented model for News20}
	\label{fig:data_reduction}
\end{figure}

We hypothesized that as Knowledge Graph is feeding more information to the model, we can achieve better performance with less training data. To verify this we have performed experiments on varying dataset fractions for 20Newsgroups dataset as shown in Figure~\ref{fig:data_reduction}. From the plot, we observe that KG augmented LSTM with 70\% data outperforms the baseline model with full dataset support, thereby reducing the dependency on labeled data by 30\%. 

\begin{figure*}[h!]
	\hspace*{1cm}\subfloat[]{\includegraphics[scale=0.8]{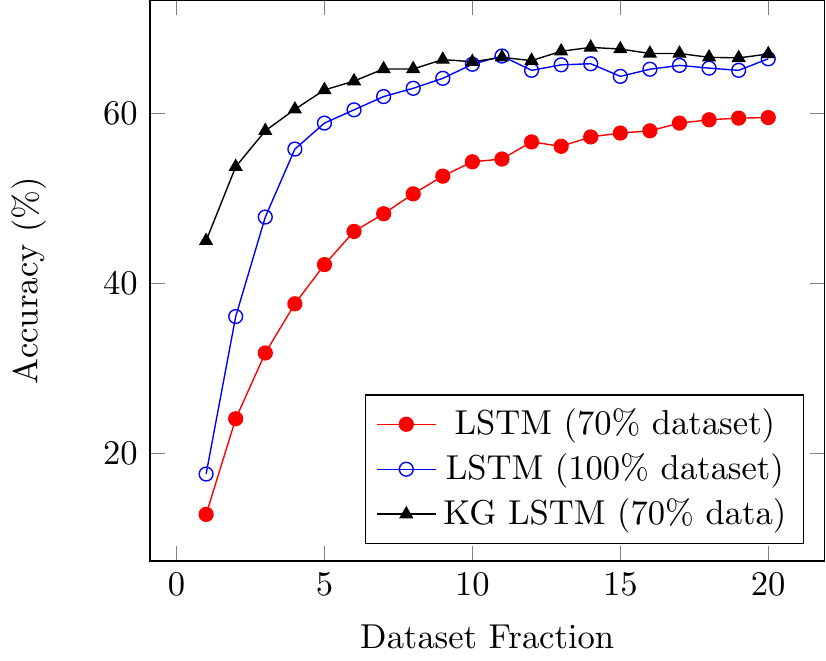}}\hspace*{.5cm}
	\subfloat[]{\includegraphics[scale=0.8]{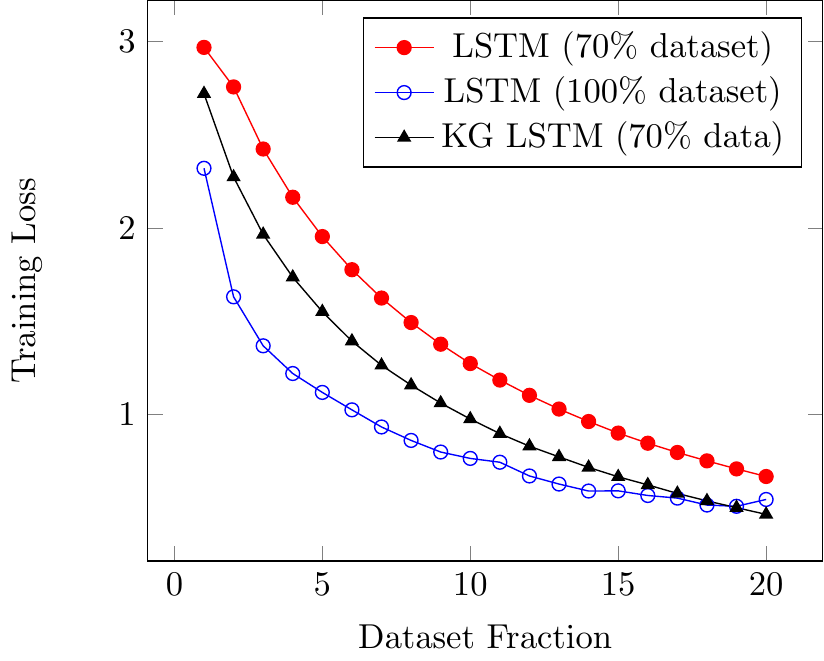}}\\
	
	\hspace*{1cm}\subfloat[]{\includegraphics[scale=0.8]{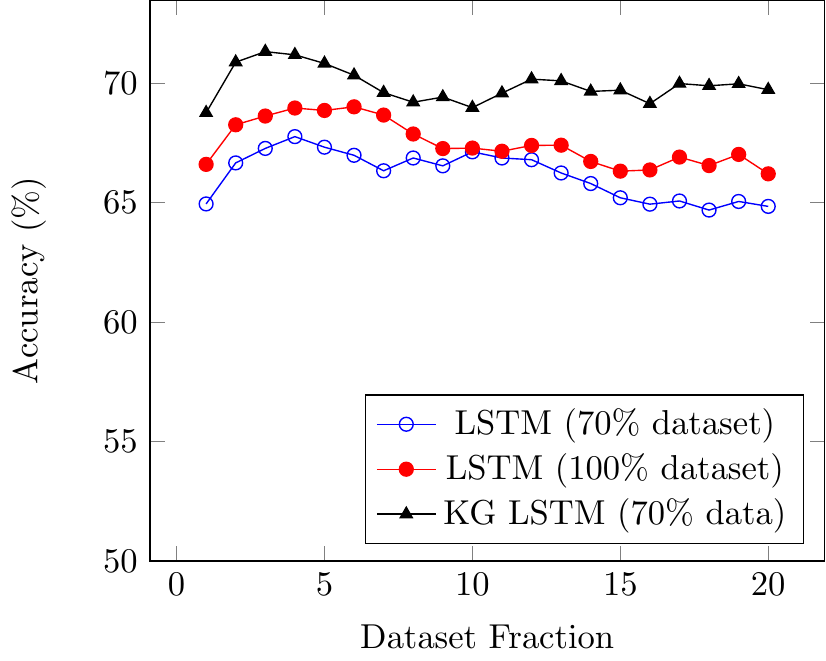}}\hspace*{.5cm}
	\subfloat[]{\includegraphics[scale=0.8]{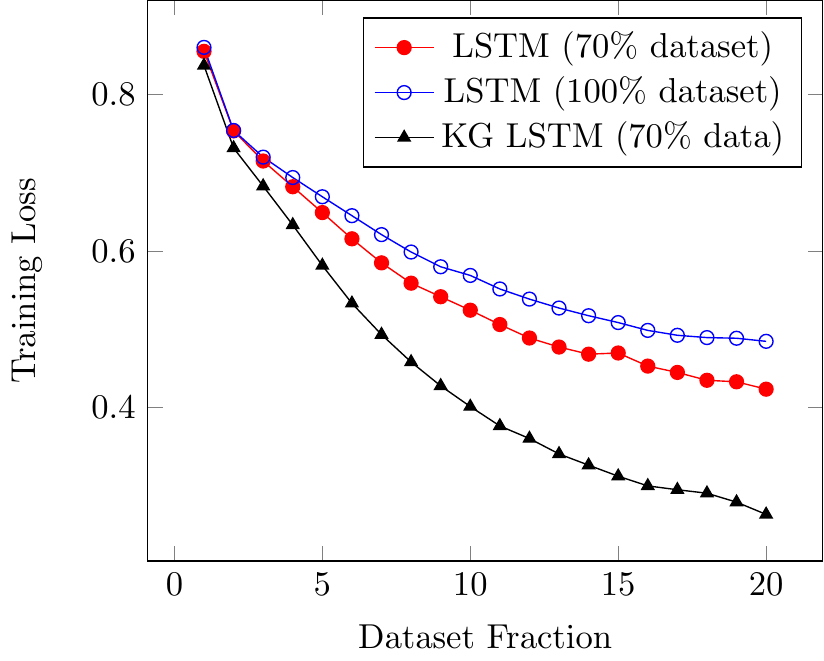}}\\
	\caption{(a) Accuracy Plot over training epochs for LSTM (using full \& 70\% dataset) and KG augmented LSTM (using 70\% dataset ) for News20 task (b) Corresponding Training Loss plots for the aforementioned methods using News20 dataset (c) Accuracy Plot over training epochs for LSTM (using full \& 70\% dataset) and KG augmented LSTM (using 70\% dataset ) for SNLI task (d) Corresponding Training Loss plots for the aforementioned methods using SNLI dataset}
	\label{fig:reduced}
\end{figure*}
We also designed an experiment to compare the accuracy of the baseline model trained on full training data and compared it with the accuracy of the KG augmented model trained with just 70\% of the training data for 20Newsgroups and SNLI datasets. The accuracy and training loss plots across training epochs is given in Figure~\ref{fig:reduced}. Even with just 70\% of the data, KG augmented model is able to achieve better accuracy compared to the vanilla LSTM model trained on the full data. 
 This clearly indicates that relevant information pertaining to the task is retrieved from the knowledge graph and the training loss reduction is not due to lesser data only. Also note that training loss is substantially less for KG LSTM compared to normal LSTM when the dataset size is reduced. This result is very promising, to reduce the large labeled training data requirement of large deep learning models, which is hard to come by.

\section{Relevant Previous Work}
\label{secprevious}

The basic idea of infusing general world knowledge for learning tasks, especially for natural language processing, has not been attempted before. For multi-label image classification, the use of KGs has been pursued recently by ~\cite{marino2016more}. In this work, they first obtain labels of the input data (using a different model), use these labels to populate features from the KG and in turn use these features back for the final classification. For NLP tasks the information needed may not necessarily depend on the final class, and we are directly using all the information available in the input for populating the relevant information from the knowledge graphs. Our attempt is very different from Transfer Learning~\cite{pan2010survey}. In Transfer Learning the focus is on training the model for one task and tuning the trained model to use it for another task. This is heavily dependent on the alignment between source task and destination task and transferred information is in the model. In our case, general world knowledge is being infused into the learning model for any given task. By the same logic, our work is different from domain adaptation~\cite{glorot2011domain} as well. There has been attempts to use world knowledge ~\cite{song2017machine} for creating more labeled training data and providing distant supervision etc. Incorporating Inductive Biases~\cite{ridgeway2016survey} based on the known information about a domain onto the structure of the learned models, is an active area of research. However our motivation and approach is fundamentally different from these works.

\section{Conclusion \& Future Work}
\label{seccon}

In this work we have illustrated the need for incorporating world knowledge in training task specific models. 
We presented a novel convolution-based architecture to reduce the attention space over entities and relations that outperformed other models. With significant improvements over the vanilla baselines for two well known datasets, we have illustrated the efficacy of our proposed methods in enhancing the performance of deep learning models. We showcased that the proposed method can be used to reduce labeled training data requirements of deep learning models. Although in this work we focused only on NLP tasks and using LSTM as the baseline model, the proposed approach is applicable for other domain tasks as well, with more complicated deep learning models as baseline. To the best of our knowledge this is the first attempt at infusing general world knowledge for task specific training of deep learning models.

\noindent Being the first work of its kind, there is a lot of scope for improvement.  A more sophisticated model which is able to retrieve facts more efficiently from millions of entries can be formulated. Currently we have focused only on a flat attention structure, a hierarchical attention mechanism would be more suitable. The model uses soft attention to enable training by simple stochastic gradient descent. Hard attention over facts using reinforcement learning can be pursued further. This will further help in selection of multi-facts, that are not of similar type but relevant to the task. The convolution based model, helped to reduce the space over entities and relationships over which attention had to be generated. However more sophisticated techniques using similarity based search~\cite{wang2014hashing,mu2017deep} can be pursued towards this purpose. The results from the initial experiments illustrates the effectiveness of our proposed approach, advocating further investigations in these directions.

\bibliographystyle{acl}
\bibliography{KG-Attention-NAACLHLT2018}
\end{document}